\documentclass[sigconf]{acmart}

\usepackage{algorithm}
\usepackage[noend]{algpseudocode}

\makeatletter
\def\BState{\State\hskip-\ALG@thistlm}
\makeatother

\def\BibTeX{{\rm B\kern-.05em{\sc i\kern-.025em b}\kern-.08emT\kern-.1667em\lower.7ex\hbox{E}\kern-.125emX}}
    
\copyrightyear{2019}
\acmYear{2019}
\setcopyright{acmlicensed}
\acmConference[KDD '19]{KDD '19: International Workshop on Fashion and KDD}{August 2019}{Anchorage, AK}
\acmBooktitle{KDD '19: International Workshop on Fashion and KDD, August 2019, Anchorage, AK}
\acmPrice{15.00}
\acmDOI{10.1145/1122445.1122456}
\acmISBN{978-1-4503-9999-9/18/06}

\usepackage{booktabs}
\usepackage{subcaption}

\begin{document}

%
\title{Predicting Next-Season Designs on High Fashion Runway}

%
\author{Yusan Lin}
\email{yusalin@visa.com}
\affiliation{%
  \institution{Visa Research}
  \streetaddress{385 Sherman Ave}
  \city{Palo Alto}
  \state{California}
  \postcode{94306}
}

\author{Hao Yang}
\email{haoyang@visa.com}
\affiliation{%
  \institution{Visa Research}
  \streetaddress{385 Sherman Ave}
  \city{Palo Alto}
  \state{California}}

%

%
\begin{abstract}
Fashion is a large and fast-changing industry. Foreseeing the upcoming fashion trends is beneficial for fashion designers, consumers, and retailers. However, fashion trends are often perceived as unpredictable due to the enormous amount of factors involved into designers' subjectivity. In this paper, we propose a fashion trend prediction framework and design neural network models to leverage structured fashion runway show data, learn the fashion collection embedding, and further train RNN/LSTM models to capture the designers' style evolution. Our proposed framework consists of (1) a runway embedding learning model that uses fashion runway images to learn every season's collection embedding, and (2) a next-season fashion design prediction model that leverage the concept of designer style and trend to predict next-season design given designers. Through experiments on a collected dataset across 32 years of fashion shows, our framework can achieve the best performance of 78.42\% AUC on average and 95\% for an individual designer when predicting the next season's design.
\end{abstract}

%
%


%

%

%
\maketitle

\section{Introduction}

Fashion is a fast-paced and dynamic industry. While the majority of fashion consumers obtain their fashion-related products from the mass market, the trends are often driven by high-fashion designers. In the fashion industry, the high-fashion designers (e.g., Chanel and Christian Dior) are the innovators that propose new fashion design ideas, while the mass market (e.g., Nike and GAP) and fast fashion brands (e.g., Zara and H\&M) are the followers. Fashion trends have been long-perceived as a product of subjective process. However, researchers have shown that fashion trends often follow specific patterns \cite{he2016ups,crane2012fashion,sproles1981analyzing}.

Fashion trend forecasting is not a new problem. WGSN has been the long-standing brand that provides fashion forecast reports every season since 1998, with clients including Coach, Nike, Adidas, and Levis'.\footnote{\url{https://www.wgsn.com/}} However, to the best of the authors' knowledge, the reports they produce are more based on qualitative analysis and consists of suggestive insights. While these are incredibly valuable resources, we believe a more precise, large-scale, and quantitative prediction is also necessary. Google once released a report on fashion trends based on search queries in 2016.\footnote{https://www.thinkwithgoogle.com/advertising-channels/search/fashion-trends-2016-google-data-consumer-insights/} While a further pursuit of the insightful results is yet to continued since then. Being able to foresee the upcoming trends in fashion has various benefits. First of all, from the designers' perspective, knowing the competitors' potential next design can help adjust their in-house designs for the next-season collection. Secondly, from the retailers' perspective, knowing ahead what trends will take off soon helps them plan of what inventory to stock up. Thirdly, from the consumers' perspective, for those that are highly trend-aware (such as online fashion influencers), knowing the popular trends in advance help them stay on the top of the fashion game.

While the change of fashion trends may seem extremely volatile and irregular, the core that drives the change of trends in the fashion industry is heavily organized and periodic: fashion shows. Fashion shows are viewed as one of the most critical events in the fashion industry that drive the fashion trends forward every fashion season for decades. We hence focus on the fashion show information to construct our fashion trend prediction system.

But even with the fashion show information, capturing the fashion trends and being able to predict the next fashion design is challenging. First of all, how to extract meaningful information from the visual data of fashion shows can be difficult. Although there has been an abundance of deep learning models built to learn high-quality embedding of images, the datasets the models are trained on are very general and not fashion focused. Secondly, fashion designers' styles change over time. How to encounter such evolvement is a critical question. Thirdly, besides the changing styles of designers, fashion designs are also influenced by the overall industry trend. A mechanism that can capture the concept of the overall trend in the fashion industry at a specific time is needed.

In this paper, we propose to leverage the fashion show data to do \emph{next-season design prediction}. We collected fashion show data of three decades, which consists of the images of each designer's collections in each fashion show. We design a prediction framework that can achieve the following three. (1) It takes in the visual of fashion show images, learns the embedding of each fashion show. (2) It trains an RNN/LSTM model \cite{rumelhart1988learning,hochreiter1997long} for each fashion designer based on their fashion designs, along with other designs in the industry.  (3) It uses the learned fashion show embedding and trained RNN/LSTM models to predict, given a fashion designer and its past designs, the design they will put out in the next season. For predicting the next season's design through a Bayesian Personalized Ranking formulation, the highest area under curve (AUC) we achieve on average 78.42\% using LSTM, and even for an individual designer with 95\% AUC.

The rest of this paper is organized as follows. In Section \ref{sec:background}, we introduce the background and literature related to fashion shows and fashion research. We then conduct data analysis on the runway image data in Section \ref{sec:analysis}. Our proposed framework is presented in Section \ref{sec:proposed} and evaluated in Section \ref{sec:evaluation}. We finally conclude this work in Section \ref{sec:conclusion}.
\section{Fashion Trend Research} \label{sec:background}


The origin of fashion shows goes back to the 1800s in Paris, while it prevailed in the 1920s in the US among the major department stores. It was not until the 1970s that fashion designers started to hold fashion shows outside of the department stores to showcase their newly released collections.\footnote{\url{https://en.wikipedia.org/wiki/Fashion_show}} Buyers from retail stores attend the fashion shows to decide what collections to in stock to their stores for the next seasons, and make the order after the shows. Fashion journalists also attend the shows to report on the newly released collections. The designers also, through the fashion shows, learn about what other designers produce in the season. Such events are viewed as the most important factors of deciding the new fashion trends \cite{skov2006role,jackson2007process}. 

Nowadays, the fashion shows take place in every fashion seasons, spring, resort, fall, in the major fashion cities, New York, Paris, and Milan. For the fashion seasons, spring shows are held around September, fall shows are held around February, and resort are usually held during summer, in between fall and spring shows. For spring and fall shows, they can be further divided into \emph{ready-to-wear} and \emph{couture} shows. Ready-to-wear shows display the collections that will be in stores, ready for the consumers to purchase, while couture shows display the collections that are for custom-made only.

The study of fashion trends can go back to several decades ago. In the 1960s, Blumer studied the trends of fashion from a theoretical point of view, which was a popular approach to study the fashion industry back then \cite{blumer1969fashion,reynolds1968cars,bronfenbrenner1966trends}. In the 1970s, the idea of \emph{fashion leaders} and \emph{design diffusion} also started to receive attention from the fashion researchers \cite{brett1975perceptual,schrank1973correlates,summers1970identity}. Afterward, the studies of the fashion industry and trends have moved toward a more quantitative direction. Tigert et al. studied fashion evolvement using fashion buyer data \cite{tigert1976fashion}. Belleau studied the cycles of dress length changes over the decades using measurements obtained from paintings since 1860 \cite{belleau1987cyclical}.

Nowadays, with the immense amount of data accessible, researchers can study fashion through an empirical approach leveraging the rich data of fashion, including online social networks, fashion magazine archives, and fashion runway images. On the consumer end, Sanchis-Ojeda et al. studied the fashion trends through consumer click rates \cite{sanchis2016detection}, and He et al. studied the fashion trends using Amazon's recommendation dataset \cite{he2016ups}. On the designer end, Vittayakorn et al. and Furukawa et al. both explored using visuals to detect fashion trends on the runway \cite{vittayakorn2015runway,furukawa2019visualisation}. We later present the strong signals visuals can convey from the fashion runway images in Section \ref{sec:analysis}.
\begin{figure}[t]
	\centering
	\begin{subfigure}[t]{.48\linewidth}
		\centering
		\includegraphics[width=\linewidth]{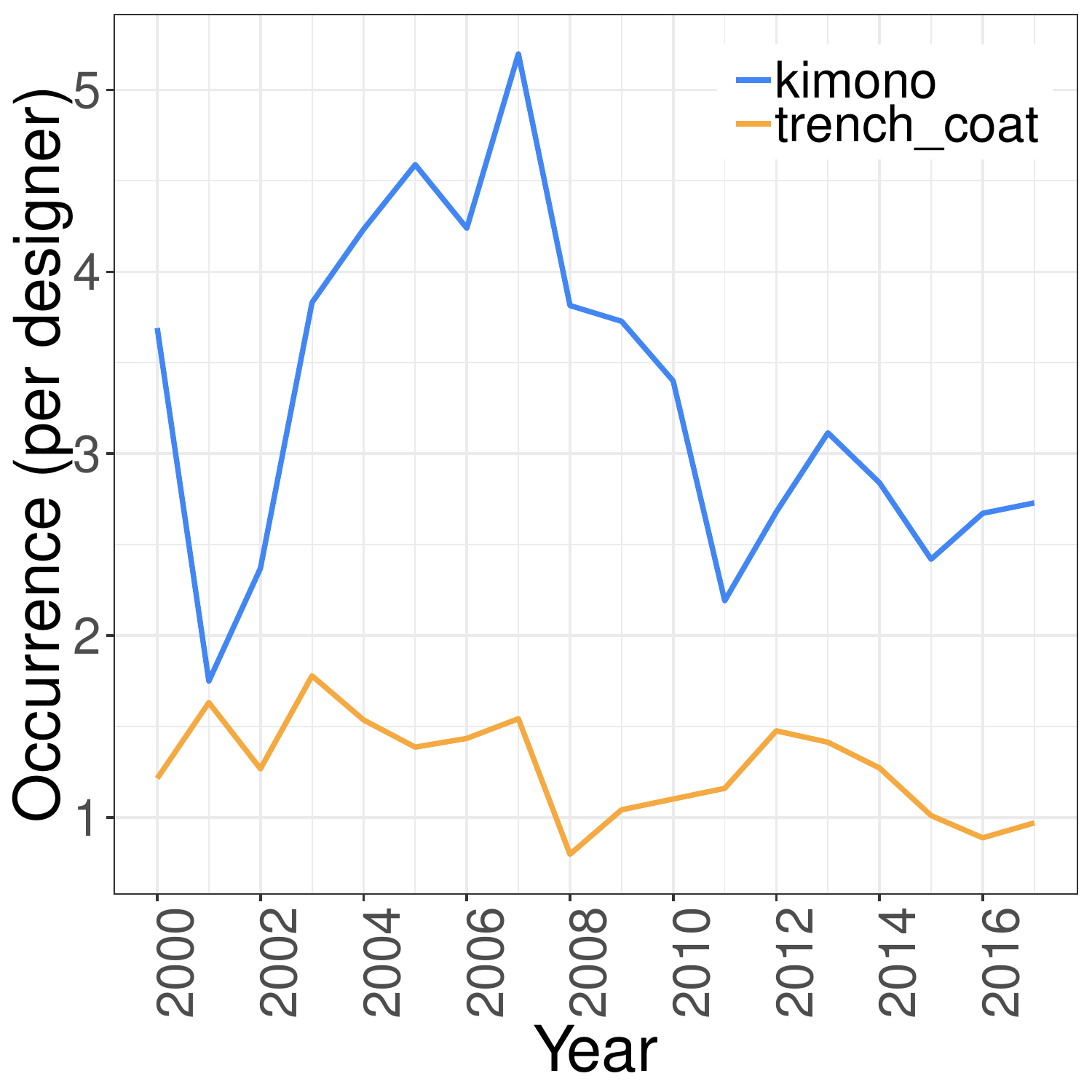}
		\caption{Kimono v.s. trench coat}
		\label{fig:kimono_vs_trenchcoat}
	\end{subfigure} \hfill
	\begin{subfigure}[t]{.48\linewidth}
		\centering
		\includegraphics[width=\linewidth]{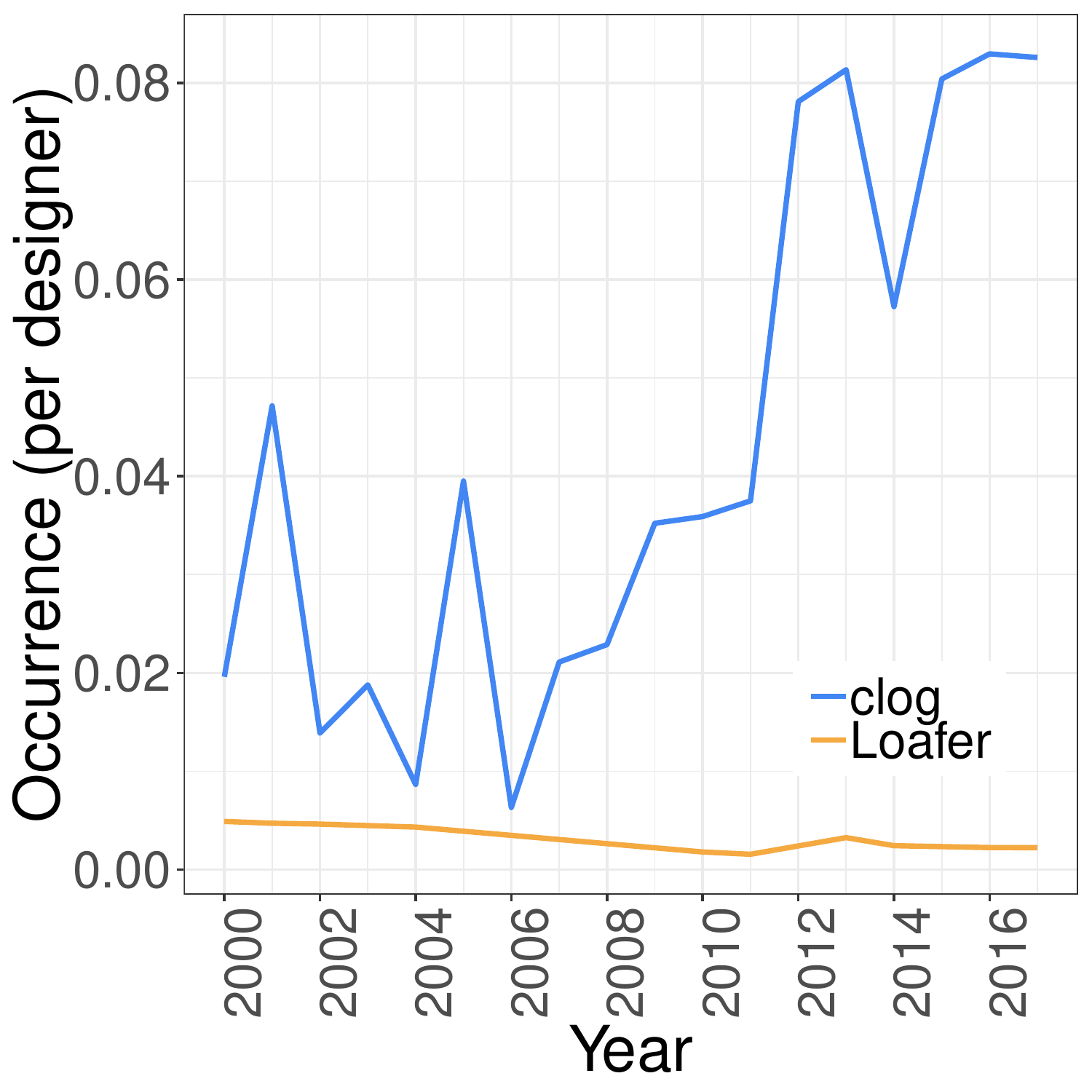}
		\caption{Loafer v.s. clog}
		\label{fig:loafer_vs_clog}
	\end{subfigure} \hfill	
	\begin{subfigure}[t]{.48\linewidth}
		\centering
		\includegraphics[width=\linewidth]{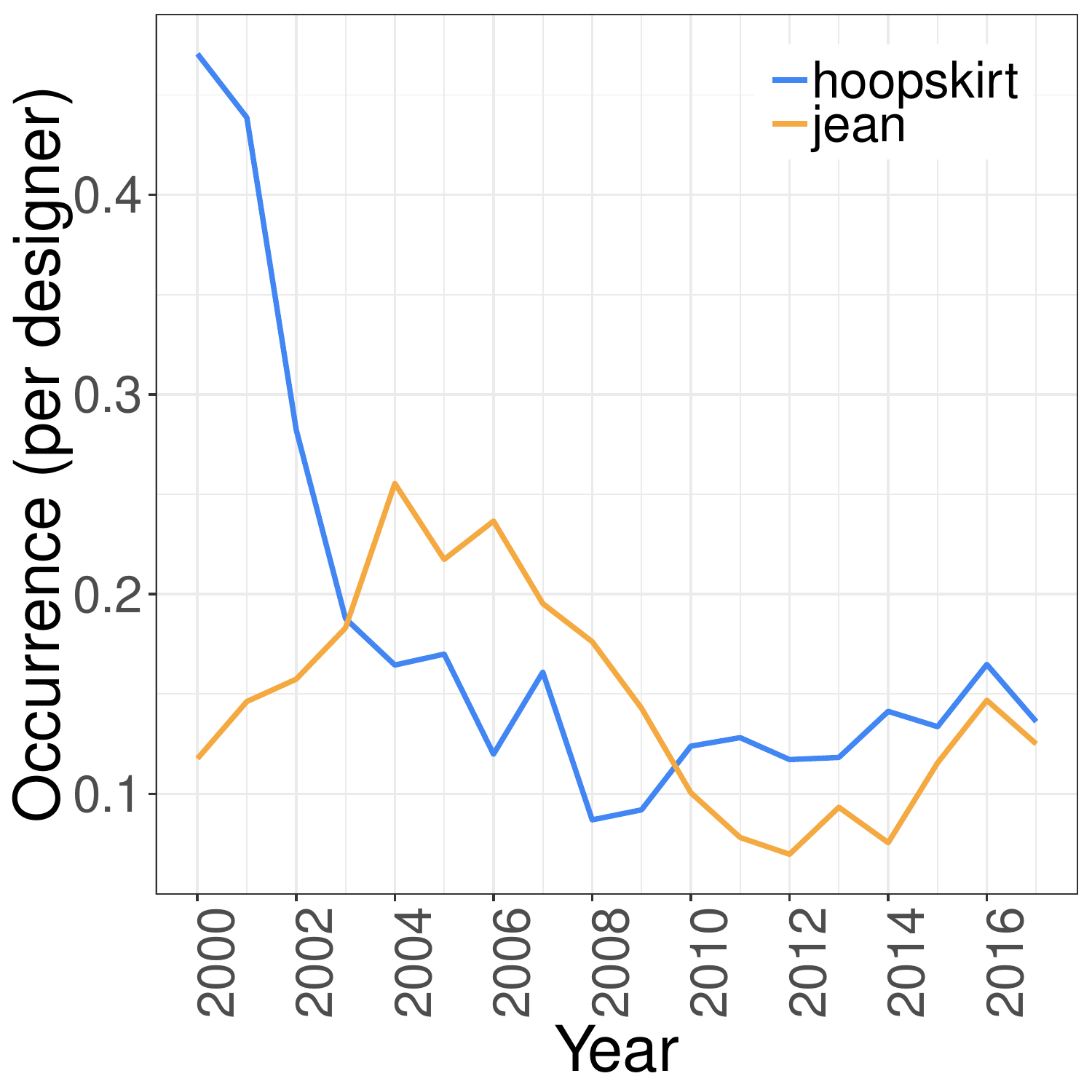}
		\caption{Hoopskirt v.s. jean}
		\label{fig:hoopskirt_vs_jean}
	\end{subfigure} \hfill
	\begin{subfigure}[t]{.48\linewidth}
		\centering
		\includegraphics[width=\linewidth]{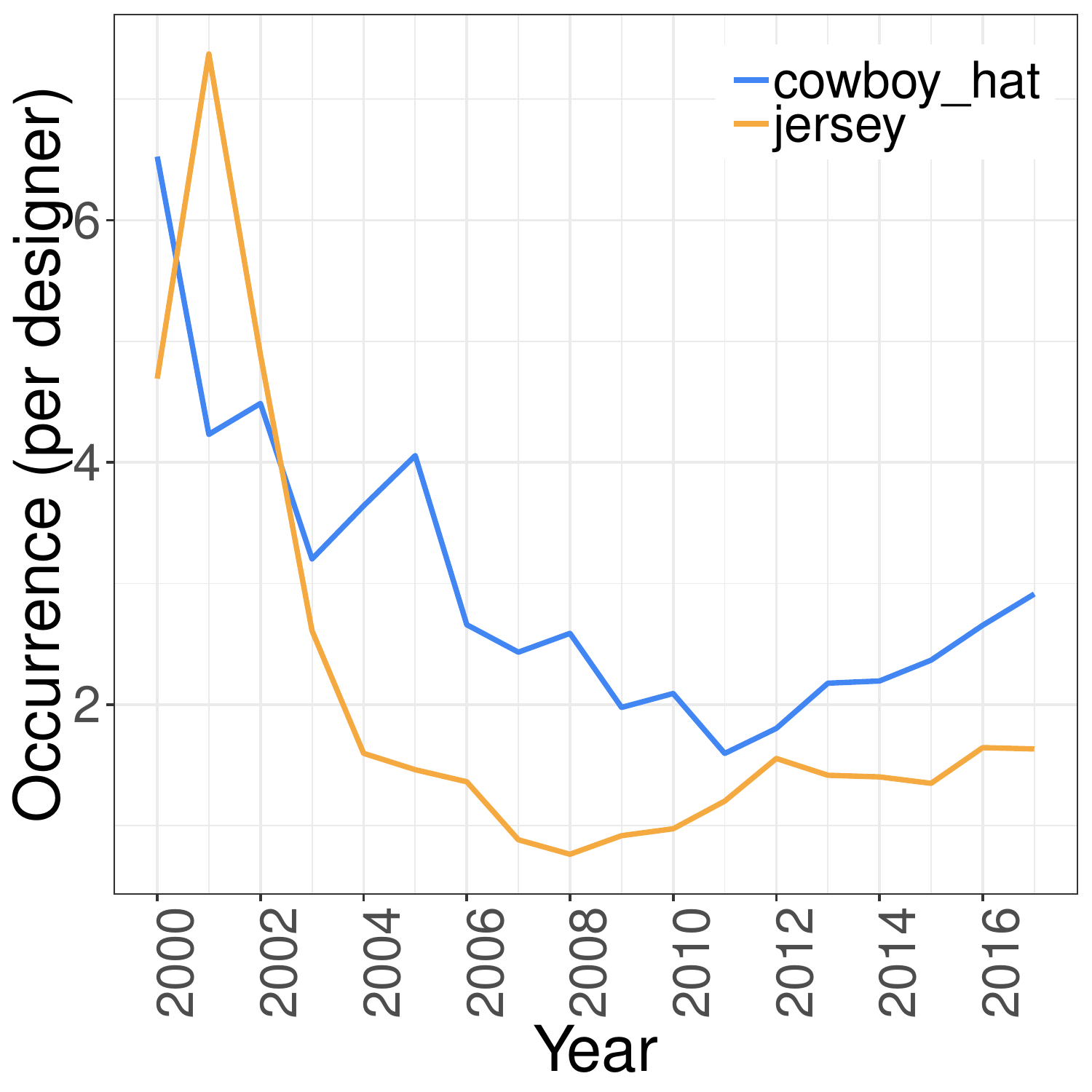}
		\caption{Cowboy hat v.s. jersey}
		\label{fig:cowboyhat_vs_jersey}
	\end{subfigure}
	\caption{Design trend detected by ImageNet pre-trained DenseNet.}
	\label{fig:imagenet_trends}
\end{figure}

\section{Runway Image Analysis} \label{sec:analysis}

Before introducing the design of our system framework, we first conduct a brief analysis on the runway show images we collected to show the rich information of fashion designs we can infer from the images using deep learning neural network models. We used 32 years of runway data, with 952 unique fashion designers, 8965 fashion shows, and 256,907 unique looks in fashion shows. Note that for this paper, we leverage only the fall and spring ready-to-wear collections since it reflects more directly to the apparels worn by consumers and timely fashion trends.

To convert runway images to more quantifiable information that we can analyze, we pass through all the collected runway images to a DenseNet that is pre-trained on ImageNet with 1000 classes \cite{DBLP:conf/cvpr/HuangLMW17,deng2009imagenet}. We retrieve the top 10 predicted classes of each image based on the classified probabilities and view them as the possible objects appeared in the images. For each class, we plot their occurrences normalized by the number of designers that had fashion shows in that year. 

Figure \ref{fig:imagenet_trends} shows the trend comparisons between different fashion designs detected by ImageNet.\footnote{We only plot trends from 2000 and onwards since the years before that include very few designers.} As shown, we can observe drastic changes in trends in the detected fashion designs. For example, as shown in Figure \ref{fig:kimono_vs_trenchcoat}, the trend of stable pieces such as trench coats does not change as much as trendy pieces such as kimonos, similar as the trends of loafers (stable piece) and clogs (trendy piece) in Figure \ref{fig:loafer_vs_clog}. Besides differentiating trendy pieces from stable pieces, Figures \ref{fig:hoopskirt_vs_jean} and \ref{fig:cowboyhat_vs_jersey} also show the changes of different fashion designs' trends. In particular, Figure \ref{fig:cowboyhat_vs_jersey} shows the declines of both cowboy hats and jersey, which were both considered as extremely trendy in the 2000s.\footnote{https://www.wmagazine.com/gallery/paris-hilton-best-2000s-style/}\footnote{https://www.buzzfeed.com/hnigatu/iconic-fashion-trends-from-the-early-2000s/}

Through the analysis of runway images using DenseNet pre-trained on ImageNet without any fine-tuning, we may confirm that pre-trained CNN models are capable of assisting us to extract meaningful visual information from the fashion runway images, even though ImageNet is not a fashion-focused dataset. Based on such confirmation, we develop our runway design prediction system relying on the visual information extracted by pre-trained CNN models.

\begin{figure}[t!]
	\centering
	\includegraphics[width=.9\linewidth]{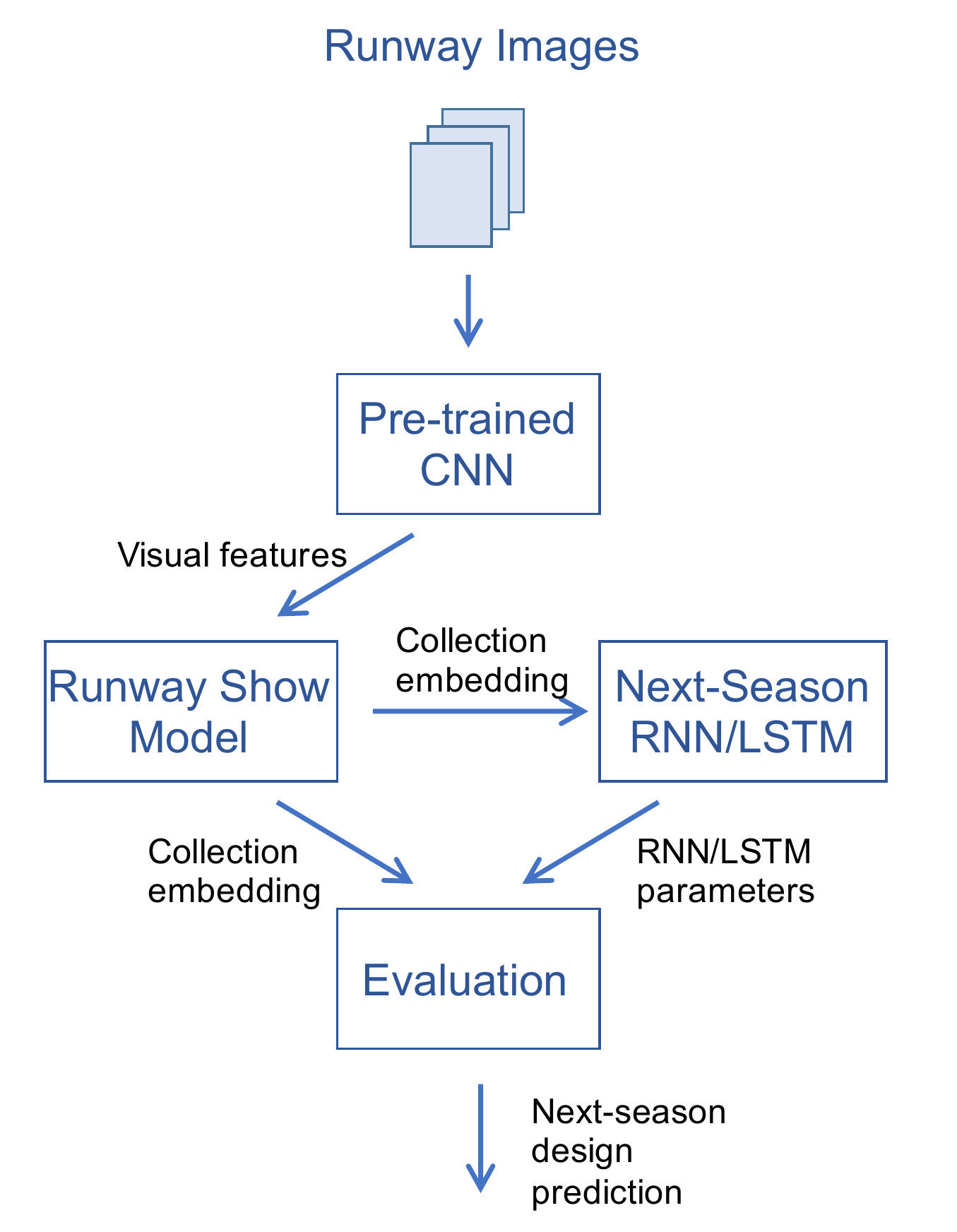}
	\caption{System framework overview.}
	\label{fig:system_framework}
\end{figure}

\section{Methodology}\label{sec:proposed}

\begin{figure*}[t!]
    \centering
	\includegraphics[width=.9\linewidth]{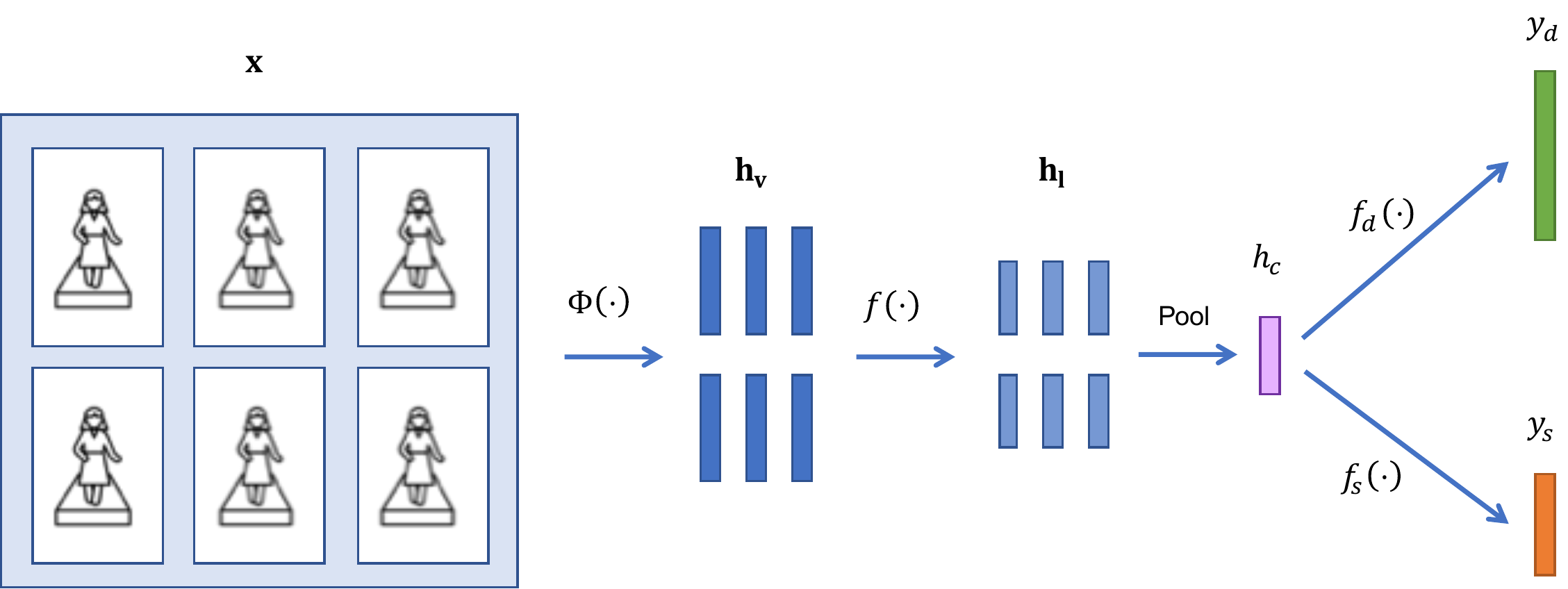}
	\caption{Runway show embedding learning model}
	\label{fig:runway_show_embedding_network}
\end{figure*}

\begin{figure}[t!]
    \centering
	\includegraphics[width=.95\linewidth]{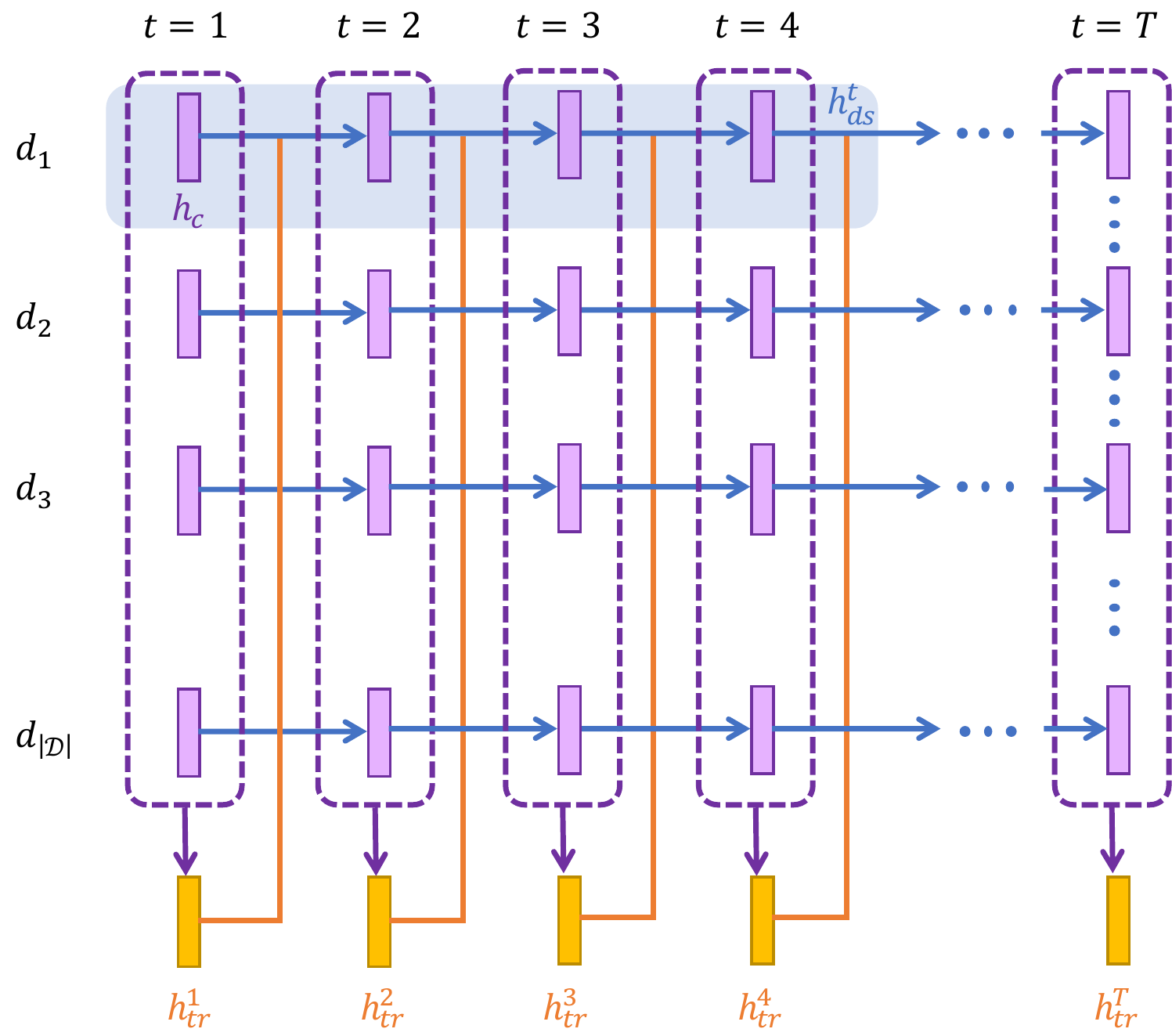}
	\caption{Next-season prediction RNN/LSTM model}
	\label{fig:next_season_lstm}
\end{figure}

The system framework we propose in this paper consists of two main modules: (1) runway show embedding learning model (Figure \ref{fig:runway_show_embedding_network}) and (2) next-season prediction RNN model (Figure \ref{fig:next_season_lstm}). Overview of the system design is shown in Figure \ref{fig:system_framework}. For clarity, we summarize the notations used in this paper in Table \ref{table:symbol_definition}. 

\begin{table}[t!]
\caption{Symbol definition}
\label{table:symbol_definition}
\begin{tabular}{@{}cl@{}}
\toprule
\textbf{Symbol} & \textbf{Definition}      \\ \midrule
$\mathcal{D}$               & Designer set             \\
$\mathcal{S}$                & Season set               \\
$\mathcal{X}$                & Look set                 \\ \midrule
$x$                          & Look image input         \\
$y_d$                        & Designer output          \\
$y_s$                        & Season output            \\ \midrule
$\Phi(\cdot)$                & CNN model                \\
$f(\cdot)$                & Fully-connected layer    \\ \midrule
$h_v$                & Visual embedding \\
$h_l$                & Look embedding           \\
$h_c$                & Collection embedding     \\
$h_{ds}$                & Designer style embedding \\
$h_{tr}$                & Trend embedding          \\ \midrule
$W, U$                  & RNN/LSTM transitional matrices \\ 
\bottomrule
\end{tabular}
\end{table}

\subsection{Problem Formulation}

Each designer $d \in \mathcal{D}$ puts out a collection of fashion designs in season $t$. Each collection consists of a set of looks $X=\lbrace x_1, ..., x_k \rbrace \subset \mathcal{X}$, where $k$ is the number of looks in the collection and $\mathcal{X}$ is the universal set of looks. The goal is given a designer $d$, its collections $[c_1,...c_{t-1}]$, predict its design at season $t$.

\subsection{Runway Show Embedding Learning Model}

Each collection by each designer consists of multiple images that capture the outfit on the runway models, which we call them \emph{looks}. The number of looks in each collection varies. To generate an embedding for a given collection, we first pass through all the looks' images $\bold{x}$ to a pre-trained CNN model of choice $\Phi(\bold{x})$ (e.g., DenseNet \cite{DBLP:conf/cvpr/HuangLMW17}) to generate visual embedding, $\bold{h_v}$. We then pass the image embedding through a fully-connected layer $f(\bold{h_a})$ to reduce their dimension, and generate \emph{look embedding}, $\bold{h_l}$. We then do a max pooling across all the look embedding to generate a \emph{collection embedding}, $h_c$. The above process can be summarized as follows. 
\begin{align}
	\bold{h_a} &= \Phi(\bold{x})  \\
	\bold{h_l} & = f(\bold{h_a})  \\
	h_c        &= \texttt{maxpool}(\bold{h_l}) 
\end{align}

A good collection embedding should be able to capture its designer and the season it belongs to. We therefore design the model to be multi-task. Firstly, with the collection embedding $h_c$, the model predicts which designer designed this collection. $h_c$ is passed through a fully-connected layer $f_d(h_c)$, then a \texttt{softmax} layer, which further outputs $\hat{y}_d$. $\hat{y}_d$ is a $|\mathcal{D}|$-dimensional vector, where the $i$\textsuperscript{th} value in $\hat{y}_d$ represents the probability of the collection being designed by the $i$\textsuperscript{th} designer in $\mathcal{D}$. The above process is summarized as follows.
\begin{align}
	\hat{y}_d = \texttt{softmax}(f_d(h_c))
\end{align}

Secondly, also with the collection embedding $h_c$, the model predicts which season\footnote{The season here we refer to is spring, resort, fall, couture, etc, regardless of the year.} this collection is released. $h_c$ is passed through a fully-connected layer $f_s(h_c)$, then a \texttt{softmax} layer, which further outputs $\hat{y}_s$. $\hat{y}_s$ is a $|\mathcal{S}|$-dimensional vector, where the $j$\textsuperscript{th} value in $\hat{y}_s$ represents the probability of the collection being released in the $j$\textsuperscript{th} season in $\mathcal{S}$. The above process is summarized as follows. 
\begin{align}
	\hat{y}_s = \texttt{softmax}(f_s(h_c))
\end{align}

The objective of the model is to minimize the following two loss functions.
\begin{align}
	\mathcal{L}_{designer} &= \frac{1}{|\mathcal{D}|}\sum H(\hat{y}_d, y_d) \\
	\mathcal{L}_{season}   &= \frac{1}{|\mathcal{D}|}\sum H(\hat{y}_s, y_s)
\end{align}
\noindent where $H$ is the cross entropy.

\subsection{Next-Season Prediction Model}

Abundant factors are influencing what a designer will design for the next season. We believe the two most important parts are what the designers have designed in the past that define their styles, and what the industry has put out as a whole in the previous seasons. We call the first component \emph{designer style} and the second component \emph{trend}. Such concept is illustrated in Figure \ref{fig:next_season_lstm}.

The trend embedding of the whole industry at a given time $j$ can be obtained through an aggregation over all the designers' collection embedding at time $t$, which is expressed as follows.
\begin{align}
	h_{tr}^t = \texttt{maxpool} (\lbrace h_c^{t(1)},...,h_c^{t(|\mathcal{D}|)} \rbrace)
\end{align}
\noindent where $h_c^{t(i)}$ denotes designer $i$'s collection embedding at time $t$.

\begin{figure}[t!]
	\centering
	\begin{subfigure}[t]{0.5\textwidth}
		\centering
		\includegraphics[width=.75\linewidth]{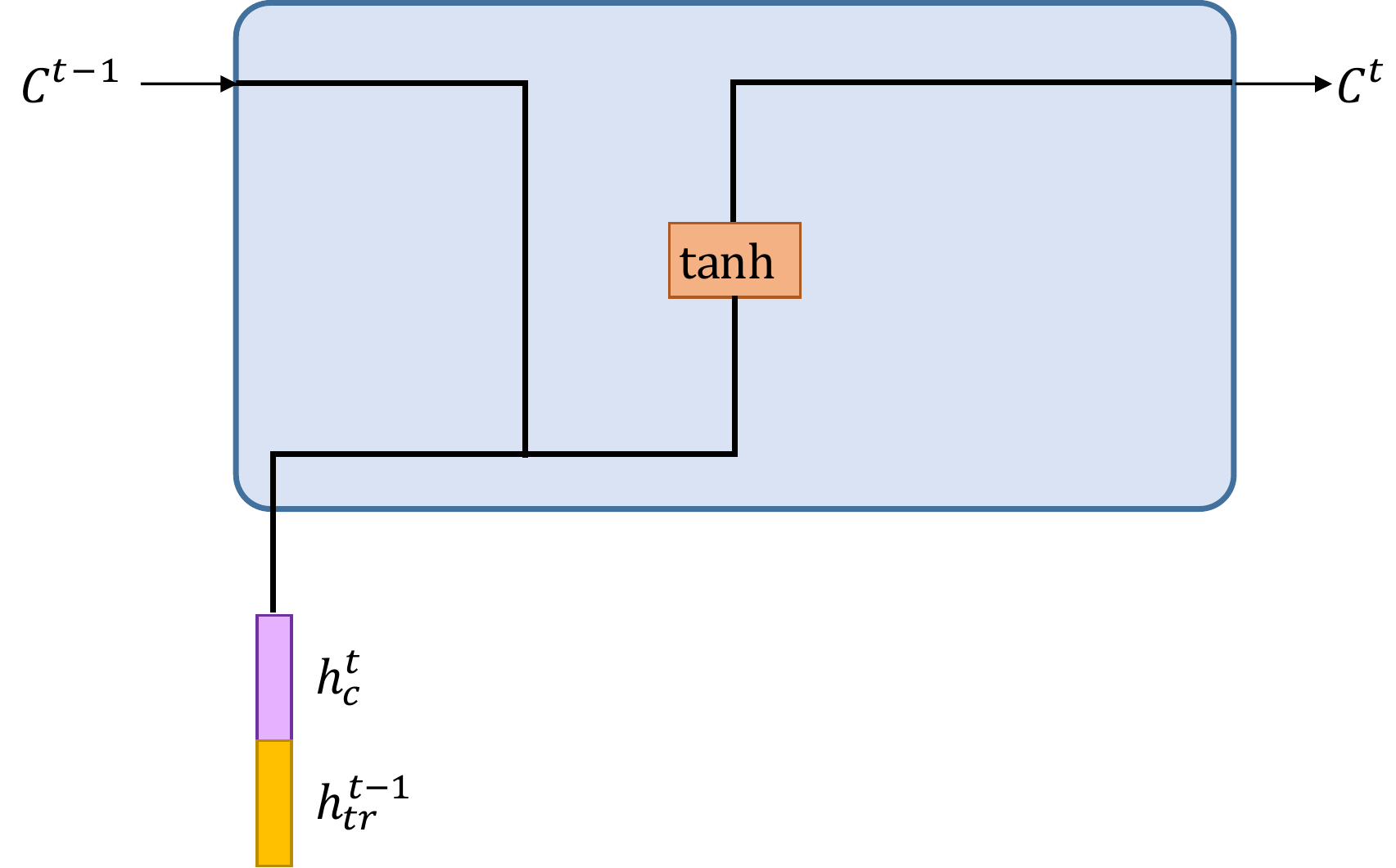}
		\caption{RNN cell.}
		\label{fig:rnn_cell}
	\end{subfigure}\vspace{5pt}
	\begin{subfigure}[t]{0.5\textwidth}
		\centering
		\includegraphics[width=.75\linewidth]{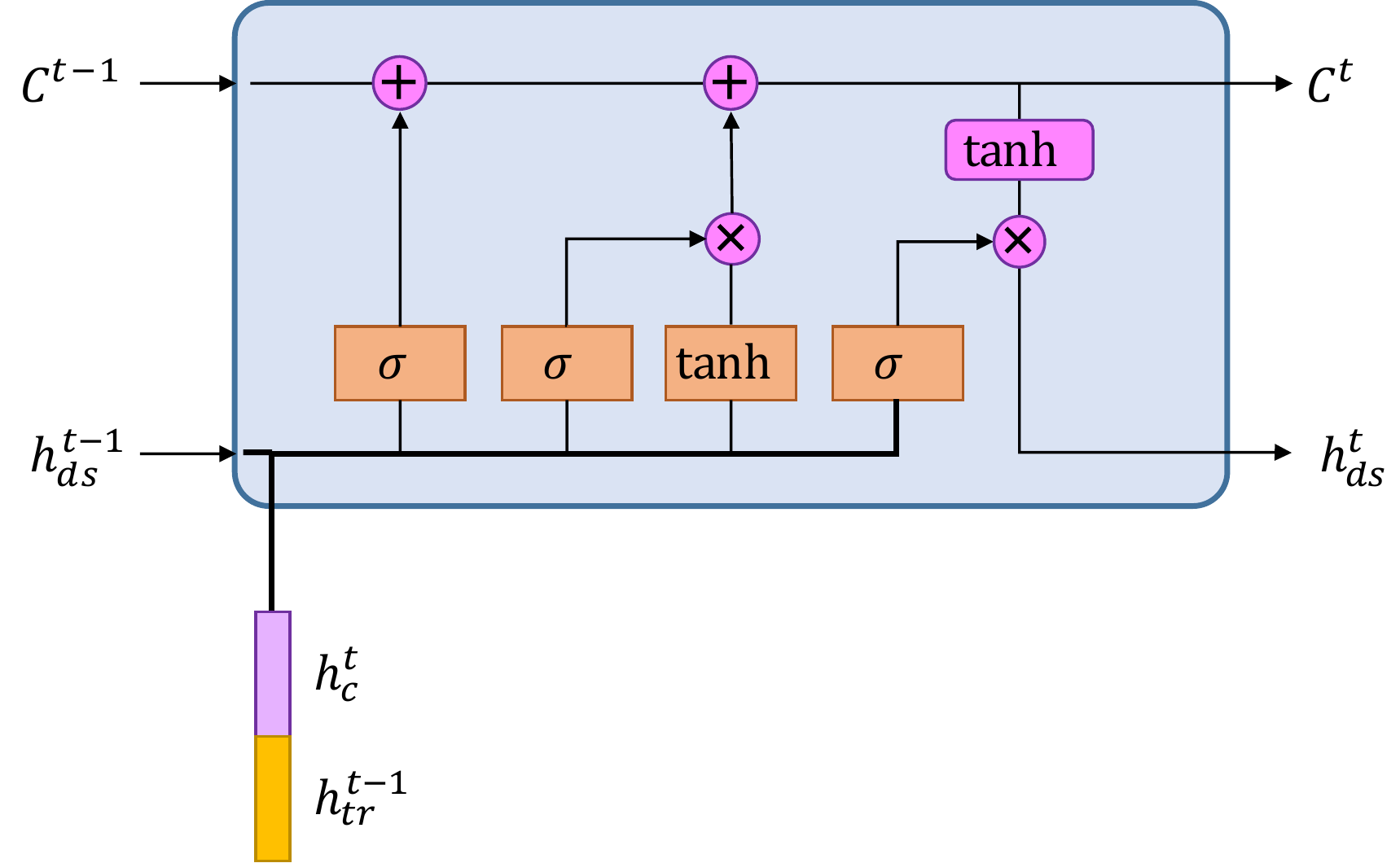}
		\caption{LSTM cell.}
		\label{fig:lstm_cell}		
	\end{subfigure}
	\caption{Cell alternatives for next-season prediction model.}
\end{figure}

The designer style of a designer $i$ at a given time $j$ can be generated using all of their collection embedding $\bold{h_c}$ from the past ($i=1...t-1$). We design two alternatives to capture such sequential evolution. The first alternative is the recurrent neural network (RNN), at time $t$, designer $i$'s designer style embedding $h_{ds}$ is obtained as follows.

\begin{align}
	h_{ds}^t = \texttt{tanh} \big( W [h_c^t || h_{tr}^{t-1}] + U h_{ds}^{t-1} +b \big)
\end{align}
\noindent where $W$ and $U$ are fully-connected layers: $W$ transforms the collection embedding $h_c$ and trend embedding $h_{tr}$ to smaller hidden embedding in RNN and $U$ transforms the design style embedding at time $t-1$ to at time $t$. By using RNN, we capture the designer's \emph{evolvement of styles} throughout the time, rather than just looking at a single snapshot. This alternative is shown in Figure \ref{fig:rnn_cell}.

Another option is to leverage long-short-term memory (LSTM) design, where the evolution of designers' styles can be modeled as below.
\begin{align}
	f_t &= \sigma_g \big( W_f [h_c^t || h_{tr}^{t-1}] + W_f h_{ds}^{t-1} + b_f \big) \\
	i_t &= \sigma_g \big( W_i [h_c^t || h_{tr}^{t-1}] + W_i h_{ds}^{t-1} + b_i \big) \\
	o_t &= \sigma_g \big( W_o [h_c^t || h_{tr}^{t-1}] + W_o h_{ds}^{t-1} + b_o \big) \\
	c_t &= f_t \odot c_{t-1} + i_t \odot \texttt{tanh} \big( W_c [h_c^t || h_{tr}^{t-1}] + U_c h_{ds}^{t-1} +b_c \big) \\
	h_t &= o_t \odot \texttt{tanh}(c_t)
\end{align}
\noindent where the purposes of $W$ and $U$ are similar as in RNN. Also, forget gates are included to enable the model's ability to capture designers' short-term and long-term dependencies on the designs and trends in the past. This alternative is shown in Figure \ref{fig:lstm_cell}.

Given a designer, a sequence of collection embedding from time $1$ to $t-1$, the objective of the model is to minimize the following loss.
\begin{align}
	\mathcal{L}_{rnn} = \frac{1}{|\mathcal{D}|}\sum\texttt{cosine} (h_c^t, \hat{h}_c^t)
\end{align}
\noindent where $\texttt{cosine}$ is the cosine distance.

\section{Evaluation} \label{sec:evaluation}

In this section, we introduce the dataset we collected and used for evaluation. We then describe our experiment setting. We finally discuss the prediction task formulation and report the results.

\subsection{Experiment Setting}

We implemented all of our models in Tensorflow. The images are passed through DenseNet to obtain image features. The image features generated by DenseNet are of dimension 50176. The look embedding $h_l$ and collection embedding $h_c$ are of dimension 256. The weights in the models are initialized using Xavier initializer and biases are initialized as zeroes. 

As a preliminary experiment, due to the cold-start problem, we focus our evaluation on the 202 designers with most fashion shows reported in our dataset.


The training of the system consists of two stages: we first train the runway show embedding model, then use the trained model to generate runway show embedding, which in turns is fed as input to the next-season prediction RNN model. We explain the training process below and summarize it in Algorithm \ref{alg:training}.

\begin{algorithm}[t!]
\caption{Model training process of runway fashion design prediction.}\label{alg:training}
\begin{algorithmic}[1]
\Procedure{TrainRunwayShowEmbedding}{}
\State $epoch \gets 0$
\While{not converged or $epoch < $ \texttt{MAX\_EPOCH}}
	\State \text{Pass batch of $(\lbrace x \rbrace, y_d, y_s)$ to }\texttt{RunwayShowEmbedding}
	\If{epoch \texttt{mod} $2 == 0$ }
		\State Optimize $\mathcal{L}_{season}$
	\Else
		\State Optimize $\mathcal{L}_{designer}$
	\EndIf
	
	\State $epoch \gets epoch + 1$
	
	\Return Learned parameters $\Theta_{runway}$
	
\EndWhile
\EndProcedure

\Procedure{TrainNextSeasonPredictionRNN/LSTM}{}
	\State $\Theta_{runway} \gets$ \texttt{TrainRunwayShowEmbedding}
	\State $h_c \gets$ Generate all collection embedding with $\Theta_{runway}$
	\State $h_{tr} \gets$ Generate all trend embedding with $\Theta_{runway}$
	\For{designer $d \in \mathcal{D}$}
		\State $epoch \gets 0$
		\While{not converged or $epoch <$ \texttt{MAX\_EPOCH}}
			\State Pass one batch of ($\Theta_{runway}, d$) \texttt{NextSeasonRNN} 
		\EndWhile
		\State Add $\Theta_{rnn}^d$ to $\Theta_{rnn}$
	\EndFor
	\Return Learned parameters $\Theta_{rnn}$
\EndProcedure

\end{algorithmic}
\end{algorithm}

\subsubsection{Runway Show Embedding Model}

Runway show embedding model is a joint-task neural network with two objective functions to minimize. Since $\mathcal{L}_{designer}$ is a multi-class classification loss (202 classes in the experiment) and $\mathcal{L}_{season}$ is a binary classification loss, the scales of the two losses are very different. Minimizing the two by linearly adding them together will dilute one of the loss' value and affect the optimization. To prevent this, we train the two objectives in each epoch interchangeably.

Each input for runway show embedding model follows the format of $(\lbrace x \rbrace, y_d, y_s)$, where the first element is a set of images. We use 70\% of the data for training, 20\% for validation and 10\% for testing. We set the batch size to 16, and use AdamDelta optimizer for backpropagation, and terminate training until both $\mathcal{L}_{designer}$ and $\mathcal{L}_{season}$ converge. For interpretability, we evaluate the model's performance on embedding learning using accuracy (i.e., the number of instances classified correctly) instead of cross entropy. At this preliminary stage, we find that given a set of runway images, the model can better distinguish the designers who create the collection (2.5\% for a 202-way classification [baseline: 0.4\%]) than the season the collection is released (50.25\% for a binary classification [baseline: 50\%]).

\subsubsection{Next-Season Prediction RNN/LSTM Model}

After training of the runway show embedding model is done, we use the trained model to create each fashion show's \emph{collection embedding}, $h_c$, and each season's \emph{trend embedding}, $h_{tr}$.

Each input for next-season prediction RNN model for a designer $d$ follows the format of $[(h_c^1, h_{tr}^1), (h_c^2, h_{tr}^2), ..., (h_c^{T_d}, h_{tr}^{T_d})]$, where $T_d$ is the maximum timestamp for designer $d$. We train a next-season prediction RNN/LSTM model for each designer with batch size 16. We use Adam optimizer with learning rate 0.0001 for backpropagation. The training stops until $\mathcal{L}_{rnn}$ converges or until the maximum number of epochs achieves (500).

\begin{figure}[t!]
	\centering
	\includegraphics[width=\linewidth]{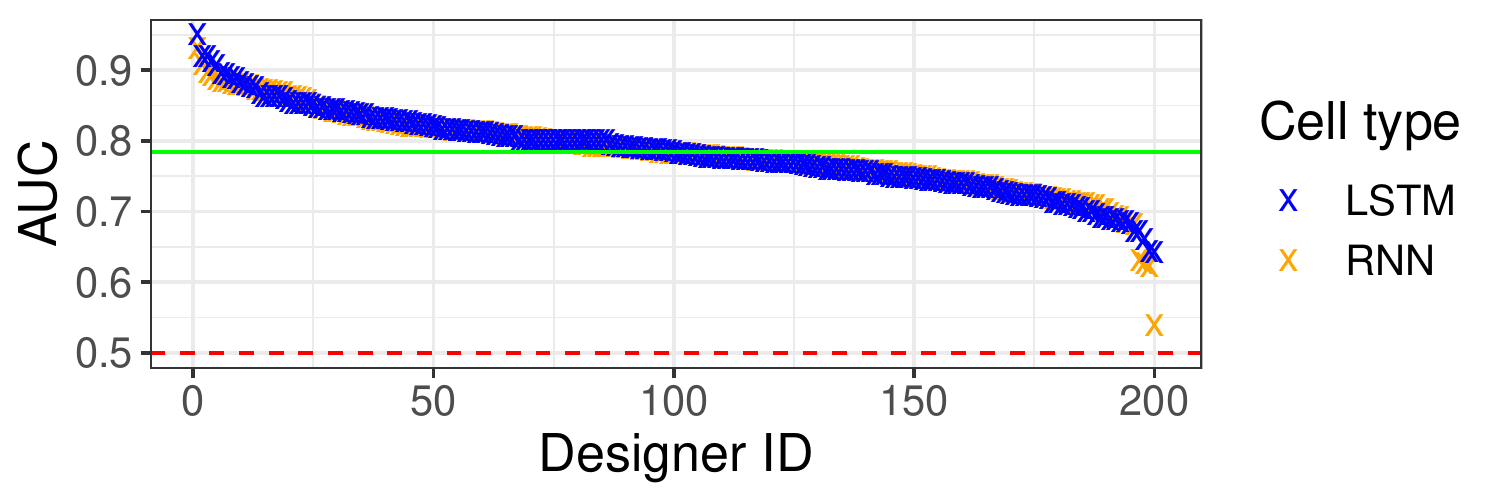}
	\caption{Individual performance on next-season prediction.}
	\label{fig:evaluation_plot}
\end{figure}

\begin{table}[t!]
\caption{Performance summary on next-season prediction.}
\label{table:evaluation_results}
\begin{tabular}{@{}llll@{}}
\toprule
\textbf{Cell type} & \textbf{Min. AUC} & \textbf{Avg. AUC} & \textbf{Max. AUC} \\ \midrule
\textbf{RNN}       & 53.85\%           & 78.40\%           & 93.02\%           \\
\textbf{LSTM}      & 64.17\%           & 78.42\%           & 95.00\%           \\
\textbf{Random}    & 50.00\%           & 50.00\%           & 50.00\%           \\ \bottomrule
\end{tabular}
\end{table}

\subsection{Next-Season Prediction Task}

To evaluate the performance of our proposed framework, we conduct a next-season design prediction task. We formulate the prediction task as follows. Given a designer $d$, her collections from time $1$ to $t-1$, her collection at time $t$ as positive collection $X_i$ and a random collection not designed by $d$ as negative collection $X_j$, the objective is to predict next-season scores so that $z^{(t-1)}_{d,i} > z^{(t-1)}_{d,j}$, where $z$ is computed as follows.
\begin{align}
	z^{(t-1)}_{d,i} = \texttt{cosine}([c_1, ..., c_{t-1}], X_i)
\end{align}

For each designer $d$ in each season, we randomly select a collection from any designer $d'$ ($d' \neq d$) in any season $s'$ as negative collection $X_j$ to form evaluation samples $\mathcal{E}_d$. We evaluate the prediction task by calculating the Area Under Curve (AUC) as follows.
\begin{align}
	AUC = \frac{1}{|\mathcal{D}|}\sum_{d\in\mathcal{D}}\frac{1}{|\mathcal{E}_d|} \sum_{(i,j) \in \mathcal{E}_d} \mathbb{I} (z^{(t-1)}_{d,i} > z^{(t-1)}_{d,j})
\end{align}
\noindent where $\mathbb{I}(\cdot)$ is an identity function counting the number of times $z^{(t-1)}_{d,i} > z^{(t-1)}_{d,j}$ is true.

We experimented with both cell alternatives, RNN and LSTM. The results are shown in Figure \ref{fig:evaluation_plot} and Table \ref{table:evaluation_results}. At this preliminary stage, without much parameter tuning, for RNN, we achieve an average AUC of 78.40\% for all the 202 designers (solid green line), which is superior to a baseline of 50\% (red dashed line) for binary classification problems. The highest performance we get for an individual designer is 93.02\% and the worst performance being 53\%. As for LSTM, we achieve an average of 78.42\% AUC, with the best performance of 95\% and worst being 64.17\%.

\section{Conclusion}\label{sec:conclusion}

In this paper, we propose a framework that leverages three decades of fashion runway image data to predict next season's fashion designs. Our framework consists of two neural networks: a runway embedding learning model and a next-season design RNN/LSTM model. We show that when compared with random guess our framework can well predict between which design will be released.

\bibliographystyle{ACM-Reference-Format}


\begin{thebibliography}{20}


\ifx \showCODEN    \undefined \def \showCODEN     #1{\unskip}     \fi
\ifx \showDOI      \undefined \def \showDOI       #1{#1}\fi
\ifx \showISBNx    \undefined \def \showISBNx     #1{\unskip}     \fi
\ifx \showISBNxiii \undefined \def \showISBNxiii  #1{\unskip}     \fi
\ifx \showISSN     \undefined \def \showISSN      #1{\unskip}     \fi
\ifx \showLCCN     \undefined \def \showLCCN      #1{\unskip}     \fi
\ifx \shownote     \undefined \def \shownote      #1{#1}          \fi
\ifx \showarticletitle \undefined \def \showarticletitle #1{#1}   \fi
\ifx \showURL      \undefined \def \showURL       {\relax}        \fi
\providecommand\bibfield[2]{#2}
\providecommand\bibinfo[2]{#2}
\providecommand\natexlab[1]{#1}
\providecommand\showeprint[2][]{arXiv:#2}

\bibitem[\protect\citeauthoryear{Belleau}{Belleau}{1987}]%
        {belleau1987cyclical}
\bibfield{author}{\bibinfo{person}{Bonnie~D Belleau}.}
  \bibinfo{year}{1987}\natexlab{}.
\newblock \showarticletitle{Cyclical fashion movement: Women's day dresses:
  1860-1980}.
\newblock \bibinfo{journal}{\emph{Clothing and textiles research journal}}
  \bibinfo{volume}{5}, \bibinfo{number}{2} (\bibinfo{year}{1987}),
  \bibinfo{pages}{15--20}.
\newblock


\bibitem[\protect\citeauthoryear{Blumer}{Blumer}{1969}]%
        {blumer1969fashion}
\bibfield{author}{\bibinfo{person}{Herbert Blumer}.}
  \bibinfo{year}{1969}\natexlab{}.
\newblock \showarticletitle{Fashion: From class differentiation to collective
  selection}.
\newblock \bibinfo{journal}{\emph{The Sociological Quarterly}}
  \bibinfo{volume}{10}, \bibinfo{number}{3} (\bibinfo{year}{1969}),
  \bibinfo{pages}{275--291}.
\newblock


\bibitem[\protect\citeauthoryear{Brett and Kernaleguen}{Brett and
  Kernaleguen}{1975}]%
        {brett1975perceptual}
\bibfield{author}{\bibinfo{person}{Joyce~E Brett} {and} \bibinfo{person}{Anne
  Kernaleguen}.} \bibinfo{year}{1975}\natexlab{}.
\newblock \showarticletitle{Perceptual and personality variables related to
  opinion leadership in fashion}.
\newblock \bibinfo{journal}{\emph{Perceptual and motor skills}}
  \bibinfo{volume}{40}, \bibinfo{number}{3} (\bibinfo{year}{1975}),
  \bibinfo{pages}{775--779}.
\newblock


\bibitem[\protect\citeauthoryear{Bronfenbrenner}{Bronfenbrenner}{1966}]%
        {bronfenbrenner1966trends}
\bibfield{author}{\bibinfo{person}{Martin Bronfenbrenner}.}
  \bibinfo{year}{1966}\natexlab{}.
\newblock \showarticletitle{Trends, cycles, and fads in economic writing}.
\newblock \bibinfo{journal}{\emph{The American Economic Review}}
  \bibinfo{volume}{56}, \bibinfo{number}{1/2} (\bibinfo{year}{1966}),
  \bibinfo{pages}{538--552}.
\newblock


\bibitem[\protect\citeauthoryear{Crane}{Crane}{2012}]%
        {crane2012fashion}
\bibfield{author}{\bibinfo{person}{Diana Crane}.}
  \bibinfo{year}{2012}\natexlab{}.
\newblock \bibinfo{booktitle}{\emph{Fashion and its social agendas: Class,
  gender, and identity in clothing}}.
\newblock \bibinfo{publisher}{University of Chicago Press}.
\newblock


\bibitem[\protect\citeauthoryear{Deng, Dong, Socher, Li, Li, and Fei-Fei}{Deng
  et~al\mbox{.}}{2009}]%
        {deng2009imagenet}
\bibfield{author}{\bibinfo{person}{Jia Deng}, \bibinfo{person}{Wei Dong},
  \bibinfo{person}{Richard Socher}, \bibinfo{person}{Li-Jia Li},
  \bibinfo{person}{Kai Li}, {and} \bibinfo{person}{Li Fei-Fei}.}
  \bibinfo{year}{2009}\natexlab{}.
\newblock \showarticletitle{Imagenet: A large-scale hierarchical image
  database}. In \bibinfo{booktitle}{\emph{2009 IEEE conference on computer
  vision and pattern recognition}}. Ieee, \bibinfo{pages}{248--255}.
\newblock


\bibitem[\protect\citeauthoryear{Furukawa, Miura, Mori, Uchida, and
  Hasegawa}{Furukawa et~al\mbox{.}}{2019}]%
        {furukawa2019visualisation}
\bibfield{author}{\bibinfo{person}{Takao Furukawa}, \bibinfo{person}{Chikako
  Miura}, \bibinfo{person}{Kaoru Mori}, \bibinfo{person}{Sou Uchida}, {and}
  \bibinfo{person}{Makoto Hasegawa}.} \bibinfo{year}{2019}\natexlab{}.
\newblock \showarticletitle{Visualisation for analysing evolutionary dynamics
  of fashion trends}.
\newblock \bibinfo{journal}{\emph{International Journal of Fashion Design,
  Technology and Education}} (\bibinfo{year}{2019}), \bibinfo{pages}{1--13}.
\newblock


\bibitem[\protect\citeauthoryear{He and McAuley}{He and McAuley}{2016}]%
        {he2016ups}
\bibfield{author}{\bibinfo{person}{Ruining He} {and} \bibinfo{person}{Julian
  McAuley}.} \bibinfo{year}{2016}\natexlab{}.
\newblock \showarticletitle{Ups and downs: Modeling the visual evolution of
  fashion trends with one-class collaborative filtering}. In
  \bibinfo{booktitle}{\emph{proceedings of the 25th international conference on
  world wide web}}. International World Wide Web Conferences Steering
  Committee, \bibinfo{pages}{507--517}.
\newblock


\bibitem[\protect\citeauthoryear{Hochreiter and Schmidhuber}{Hochreiter and
  Schmidhuber}{1997}]%
        {hochreiter1997long}
\bibfield{author}{\bibinfo{person}{Sepp Hochreiter} {and}
  \bibinfo{person}{J{\"u}rgen Schmidhuber}.} \bibinfo{year}{1997}\natexlab{}.
\newblock \showarticletitle{Long short-term memory}.
\newblock \bibinfo{journal}{\emph{Neural computation}} \bibinfo{volume}{9},
  \bibinfo{number}{8} (\bibinfo{year}{1997}), \bibinfo{pages}{1735--1780}.
\newblock


\bibitem[\protect\citeauthoryear{Huang, Liu, van~der Maaten, and
  Weinberger}{Huang et~al\mbox{.}}{2017}]%
        {DBLP:conf/cvpr/HuangLMW17}
\bibfield{author}{\bibinfo{person}{Gao Huang}, \bibinfo{person}{Zhuang Liu},
  \bibinfo{person}{Laurens van~der Maaten}, {and} \bibinfo{person}{Kilian~Q.
  Weinberger}.} \bibinfo{year}{2017}\natexlab{}.
\newblock \showarticletitle{Densely Connected Convolutional Networks}. In
  \bibinfo{booktitle}{\emph{2017 {IEEE} Conference on Computer Vision and
  Pattern Recognition, {CVPR} 2017, Honolulu, HI, USA, July 21-26, 2017}}.
  \bibinfo{pages}{2261--2269}.
\newblock
\urldef\tempurl%
\url{https://doi.org/10.1109/CVPR.2017.243}
\showDOI{\tempurl}


\bibitem[\protect\citeauthoryear{Jackson}{Jackson}{2007}]%
        {jackson2007process}
\bibfield{author}{\bibinfo{person}{Tim Jackson}.}
  \bibinfo{year}{2007}\natexlab{}.
\newblock \showarticletitle{The process of trend development leading to a
  fashion season}.
\newblock In \bibinfo{booktitle}{\emph{Fashion Marketing}}.
  \bibinfo{publisher}{Routledge}, \bibinfo{pages}{192--211}.
\newblock


\bibitem[\protect\citeauthoryear{Reynolds}{Reynolds}{1968}]%
        {reynolds1968cars}
\bibfield{author}{\bibinfo{person}{William~H Reynolds}.}
  \bibinfo{year}{1968}\natexlab{}.
\newblock \showarticletitle{Cars and clothing: understanding fashion trends}.
\newblock \bibinfo{journal}{\emph{Journal of Marketing}} \bibinfo{volume}{32},
  \bibinfo{number}{3} (\bibinfo{year}{1968}), \bibinfo{pages}{44--49}.
\newblock


\bibitem[\protect\citeauthoryear{Rumelhart, Hinton, Williams,
  et~al\mbox{.}}{Rumelhart et~al\mbox{.}}{1988}]%
        {rumelhart1988learning}
\bibfield{author}{\bibinfo{person}{David~E Rumelhart},
  \bibinfo{person}{Geoffrey~E Hinton}, \bibinfo{person}{Ronald~J Williams},
  {et~al\mbox{.}}} \bibinfo{year}{1988}\natexlab{}.
\newblock \showarticletitle{Learning representations by back-propagating
  errors}.
\newblock \bibinfo{journal}{\emph{Cognitive modeling}} \bibinfo{volume}{5},
  \bibinfo{number}{3} (\bibinfo{year}{1988}), \bibinfo{pages}{1}.
\newblock


\bibitem[\protect\citeauthoryear{Sanchis-Ojeda, Sibley, and
  Massimi}{Sanchis-Ojeda et~al\mbox{.}}{2016}]%
        {sanchis2016detection}
\bibfield{author}{\bibinfo{person}{Roberto Sanchis-Ojeda},
  \bibinfo{person}{Daragh Sibley}, {and} \bibinfo{person}{Paolo Massimi}.}
  \bibinfo{year}{2016}\natexlab{}.
\newblock \showarticletitle{Detection of fashion trends and seasonal cycles
  through the analysis of implicit and explicit client feedback}. In
  \bibinfo{booktitle}{\emph{KDD Fashion Workshop}}.
\newblock


\bibitem[\protect\citeauthoryear{Schrank and Lois~Gilmore}{Schrank and
  Lois~Gilmore}{1973}]%
        {schrank1973correlates}
\bibfield{author}{\bibinfo{person}{Holly~L Schrank} {and} \bibinfo{person}{D
  Lois~Gilmore}.} \bibinfo{year}{1973}\natexlab{}.
\newblock \showarticletitle{Correlates of fashion leadership: Implications for
  fashion process theory}.
\newblock \bibinfo{journal}{\emph{Sociological Quarterly}}
  \bibinfo{volume}{14}, \bibinfo{number}{4} (\bibinfo{year}{1973}),
  \bibinfo{pages}{534--543}.
\newblock


\bibitem[\protect\citeauthoryear{Skov}{Skov}{2006}]%
        {skov2006role}
\bibfield{author}{\bibinfo{person}{Lise Skov}.}
  \bibinfo{year}{2006}\natexlab{}.
\newblock \showarticletitle{The role of trade fairs in the global fashion
  business}.
\newblock \bibinfo{journal}{\emph{Current Sociology}} \bibinfo{volume}{54},
  \bibinfo{number}{5} (\bibinfo{year}{2006}), \bibinfo{pages}{764--783}.
\newblock


\bibitem[\protect\citeauthoryear{Sproles}{Sproles}{1981}]%
        {sproles1981analyzing}
\bibfield{author}{\bibinfo{person}{George~B Sproles}.}
  \bibinfo{year}{1981}\natexlab{}.
\newblock \showarticletitle{Analyzing fashion life cycles—Principles and
  perspectives}.
\newblock \bibinfo{journal}{\emph{Journal of Marketing}} \bibinfo{volume}{45},
  \bibinfo{number}{4} (\bibinfo{year}{1981}), \bibinfo{pages}{116--124}.
\newblock


\bibitem[\protect\citeauthoryear{Summers}{Summers}{1970}]%
        {summers1970identity}
\bibfield{author}{\bibinfo{person}{John~O Summers}.}
  \bibinfo{year}{1970}\natexlab{}.
\newblock \showarticletitle{The identity of women's clothing fashion opinion
  leaders}.
\newblock \bibinfo{journal}{\emph{Journal of Marketing Research}}
  \bibinfo{volume}{7}, \bibinfo{number}{2} (\bibinfo{year}{1970}),
  \bibinfo{pages}{178--185}.
\newblock


\bibitem[\protect\citeauthoryear{Tigert, Ring, and King}{Tigert
  et~al\mbox{.}}{1976}]%
        {tigert1976fashion}
\bibfield{author}{\bibinfo{person}{Douglas~J Tigert},
  \bibinfo{person}{Lawrence~J Ring}, {and} \bibinfo{person}{Charles~W King}.}
  \bibinfo{year}{1976}\natexlab{}.
\newblock \showarticletitle{Fashion involvement and buying behavior: A
  methodological study}.
\newblock \bibinfo{journal}{\emph{ACR North American Advances}}
  (\bibinfo{year}{1976}).
\newblock


\bibitem[\protect\citeauthoryear{Vittayakorn, Yamaguchi, Berg, and
  Berg}{Vittayakorn et~al\mbox{.}}{2015}]%
        {vittayakorn2015runway}
\bibfield{author}{\bibinfo{person}{Sirion Vittayakorn}, \bibinfo{person}{Kota
  Yamaguchi}, \bibinfo{person}{Alexander~C Berg}, {and}
  \bibinfo{person}{Tamara~L Berg}.} \bibinfo{year}{2015}\natexlab{}.
\newblock \showarticletitle{Runway to realway: Visual analysis of fashion}. In
  \bibinfo{booktitle}{\emph{2015 IEEE Winter Conference on Applications of
  Computer Vision}}. IEEE, \bibinfo{pages}{951--958}.
\newblock


\end{thebibliography}


\begin{thebibliography}{00}


\ifx \showCODEN    \undefined \def \showCODEN     #1{\unskip}     \fi
\ifx \showDOI      \undefined \def \showDOI       #1{#1}\fi
\ifx \showISBNx    \undefined \def \showISBNx     #1{\unskip}     \fi
\ifx \showISBNxiii \undefined \def \showISBNxiii  #1{\unskip}     \fi
\ifx \showISSN     \undefined \def \showISSN      #1{\unskip}     \fi
\ifx \showLCCN     \undefined \def \showLCCN      #1{\unskip}     \fi
\ifx \shownote     \undefined \def \shownote      #1{#1}          \fi
\ifx \showarticletitle \undefined \def \showarticletitle #1{#1}   \fi
\ifx \showURL      \undefined \def \showURL       {\relax}        \fi
\providecommand\bibfield[2]{#2}
\providecommand\bibinfo[2]{#2}
\providecommand\natexlab[1]{#1}
\providecommand\showeprint[2][]{arXiv:#2}

\bibitem[\protect\citeauthoryear{Belleau}{Belleau}{1987}]%
        {belleau1987cyclical}
\bibfield{author}{\bibinfo{person}{Bonnie~D Belleau}.}
  \bibinfo{year}{1987}\natexlab{}.
\newblock \showarticletitle{Cyclical fashion movement: Women's day dresses:
  1860-1980}.
\newblock \bibinfo{journal}{{\em Clothing and textiles research journal\/}}
  \bibinfo{volume}{5}, \bibinfo{number}{2} (\bibinfo{year}{1987}),
  \bibinfo{pages}{15--20}.
\newblock


\bibitem[\protect\citeauthoryear{Blumer}{Blumer}{1969}]%
        {blumer1969fashion}
\bibfield{author}{\bibinfo{person}{Herbert Blumer}.}
  \bibinfo{year}{1969}\natexlab{}.
\newblock \showarticletitle{Fashion: From class differentiation to collective
  selection}.
\newblock \bibinfo{journal}{{\em The Sociological Quarterly\/}}
  \bibinfo{volume}{10}, \bibinfo{number}{3} (\bibinfo{year}{1969}),
  \bibinfo{pages}{275--291}.
\newblock


\bibitem[\protect\citeauthoryear{Brett and Kernaleguen}{Brett and
  Kernaleguen}{1975}]%
        {brett1975perceptual}
\bibfield{author}{\bibinfo{person}{Joyce~E Brett} {and} \bibinfo{person}{Anne
  Kernaleguen}.} \bibinfo{year}{1975}\natexlab{}.
\newblock \showarticletitle{Perceptual and personality variables related to
  opinion leadership in fashion}.
\newblock \bibinfo{journal}{{\em Perceptual and motor skills\/}}
  \bibinfo{volume}{40}, \bibinfo{number}{3} (\bibinfo{year}{1975}),
  \bibinfo{pages}{775--779}.
\newblock


\bibitem[\protect\citeauthoryear{Bronfenbrenner}{Bronfenbrenner}{1966}]%
        {bronfenbrenner1966trends}
\bibfield{author}{\bibinfo{person}{Martin Bronfenbrenner}.}
  \bibinfo{year}{1966}\natexlab{}.
\newblock \showarticletitle{Trends, cycles, and fads in economic writing}.
\newblock \bibinfo{journal}{{\em The American Economic Review\/}}
  \bibinfo{volume}{56}, \bibinfo{number}{1/2} (\bibinfo{year}{1966}),
  \bibinfo{pages}{538--552}.
\newblock


\bibitem[\protect\citeauthoryear{Crane}{Crane}{2012}]%
        {crane2012fashion}
\bibfield{author}{\bibinfo{person}{Diana Crane}.}
  \bibinfo{year}{2012}\natexlab{}.
\newblock \bibinfo{booktitle}{{\em Fashion and its social agendas: Class,
  gender, and identity in clothing}}.
\newblock \bibinfo{publisher}{University of Chicago Press}.
\newblock


\bibitem[\protect\citeauthoryear{Deng, Dong, Socher, Li, Li, and Fei-Fei}{Deng
  et~al\mbox{.}}{2009}]%
        {deng2009imagenet}
\bibfield{author}{\bibinfo{person}{Jia Deng}, \bibinfo{person}{Wei Dong},
  \bibinfo{person}{Richard Socher}, \bibinfo{person}{Li-Jia Li},
  \bibinfo{person}{Kai Li}, {and} \bibinfo{person}{Li Fei-Fei}.}
  \bibinfo{year}{2009}\natexlab{}.
\newblock \showarticletitle{Imagenet: A large-scale hierarchical image
  database}. In \bibinfo{booktitle}{{\em 2009 IEEE conference on computer
  vision and pattern recognition}}. Ieee, \bibinfo{pages}{248--255}.
\newblock


\bibitem[\protect\citeauthoryear{Furukawa, Miura, Mori, Uchida, and
  Hasegawa}{Furukawa et~al\mbox{.}}{2019}]%
        {furukawa2019visualisation}
\bibfield{author}{\bibinfo{person}{Takao Furukawa}, \bibinfo{person}{Chikako
  Miura}, \bibinfo{person}{Kaoru Mori}, \bibinfo{person}{Sou Uchida}, {and}
  \bibinfo{person}{Makoto Hasegawa}.} \bibinfo{year}{2019}\natexlab{}.
\newblock \showarticletitle{Visualisation for analysing evolutionary dynamics
  of fashion trends}.
\newblock \bibinfo{journal}{{\em International Journal of Fashion Design,
  Technology and Education\/}} (\bibinfo{year}{2019}), \bibinfo{pages}{1--13}.
\newblock


\bibitem[\protect\citeauthoryear{He and McAuley}{He and McAuley}{2016}]%
        {he2016ups}
\bibfield{author}{\bibinfo{person}{Ruining He} {and} \bibinfo{person}{Julian
  McAuley}.} \bibinfo{year}{2016}\natexlab{}.
\newblock \showarticletitle{Ups and downs: Modeling the visual evolution of
  fashion trends with one-class collaborative filtering}. In
  \bibinfo{booktitle}{{\em proceedings of the 25th international conference on
  world wide web}}. International World Wide Web Conferences Steering
  Committee, \bibinfo{pages}{507--517}.
\newblock


\bibitem[\protect\citeauthoryear{Hochreiter and Schmidhuber}{Hochreiter and
  Schmidhuber}{1997}]%
        {hochreiter1997long}
\bibfield{author}{\bibinfo{person}{Sepp Hochreiter} {and}
  \bibinfo{person}{J{\"u}rgen Schmidhuber}.} \bibinfo{year}{1997}\natexlab{}.
\newblock \showarticletitle{Long short-term memory}.
\newblock \bibinfo{journal}{{\em Neural computation\/}} \bibinfo{volume}{9},
  \bibinfo{number}{8} (\bibinfo{year}{1997}), \bibinfo{pages}{1735--1780}.
\newblock


\bibitem[\protect\citeauthoryear{Huang, Liu, van~der Maaten, and
  Weinberger}{Huang et~al\mbox{.}}{2017}]%
        {DBLP:conf/cvpr/HuangLMW17}
\bibfield{author}{\bibinfo{person}{Gao Huang}, \bibinfo{person}{Zhuang Liu},
  \bibinfo{person}{Laurens van~der Maaten}, {and} \bibinfo{person}{Kilian~Q.
  Weinberger}.} \bibinfo{year}{2017}\natexlab{}.
\newblock \showarticletitle{Densely Connected Convolutional Networks}. In
  \bibinfo{booktitle}{{\em 2017 {IEEE} Conference on Computer Vision and
  Pattern Recognition, {CVPR} 2017, Honolulu, HI, USA, July 21-26, 2017}}.
  \bibinfo{pages}{2261--2269}.
\newblock
\showDOI{%
\url{https://doi.org/10.1109/CVPR.2017.243}}


\bibitem[\protect\citeauthoryear{Jackson}{Jackson}{2007}]%
        {jackson2007process}
\bibfield{author}{\bibinfo{person}{Tim Jackson}.}
  \bibinfo{year}{2007}\natexlab{}.
\newblock \showarticletitle{The process of trend development leading to a
  fashion season}.
\newblock In \bibinfo{booktitle}{{\em Fashion Marketing}}.
  \bibinfo{publisher}{Routledge}, \bibinfo{pages}{192--211}.
\newblock


\bibitem[\protect\citeauthoryear{Reynolds}{Reynolds}{1968}]%
        {reynolds1968cars}
\bibfield{author}{\bibinfo{person}{William~H Reynolds}.}
  \bibinfo{year}{1968}\natexlab{}.
\newblock \showarticletitle{Cars and clothing: understanding fashion trends}.
\newblock \bibinfo{journal}{{\em Journal of Marketing\/}} \bibinfo{volume}{32},
  \bibinfo{number}{3} (\bibinfo{year}{1968}), \bibinfo{pages}{44--49}.
\newblock


\bibitem[\protect\citeauthoryear{Rumelhart, Hinton, Williams,
  et~al\mbox{.}}{Rumelhart et~al\mbox{.}}{1988}]%
        {rumelhart1988learning}
\bibfield{author}{\bibinfo{person}{David~E Rumelhart},
  \bibinfo{person}{Geoffrey~E Hinton}, \bibinfo{person}{Ronald~J Williams},
  {et~al\mbox{.}}} \bibinfo{year}{1988}\natexlab{}.
\newblock \showarticletitle{Learning representations by back-propagating
  errors}.
\newblock \bibinfo{journal}{{\em Cognitive modeling\/}} \bibinfo{volume}{5},
  \bibinfo{number}{3} (\bibinfo{year}{1988}), \bibinfo{pages}{1}.
\newblock


\bibitem[\protect\citeauthoryear{Sanchis-Ojeda, Sibley, and
  Massimi}{Sanchis-Ojeda et~al\mbox{.}}{2016}]%
        {sanchis2016detection}
\bibfield{author}{\bibinfo{person}{Roberto Sanchis-Ojeda},
  \bibinfo{person}{Daragh Sibley}, {and} \bibinfo{person}{Paolo Massimi}.}
  \bibinfo{year}{2016}\natexlab{}.
\newblock \showarticletitle{Detection of fashion trends and seasonal cycles
  through the analysis of implicit and explicit client feedback}. In
  \bibinfo{booktitle}{{\em KDD Fashion Workshop}}.
\newblock


\bibitem[\protect\citeauthoryear{Schrank and Lois~Gilmore}{Schrank and
  Lois~Gilmore}{1973}]%
        {schrank1973correlates}
\bibfield{author}{\bibinfo{person}{Holly~L Schrank} {and} \bibinfo{person}{D
  Lois~Gilmore}.} \bibinfo{year}{1973}\natexlab{}.
\newblock \showarticletitle{Correlates of fashion leadership: Implications for
  fashion process theory}.
\newblock \bibinfo{journal}{{\em Sociological Quarterly\/}}
  \bibinfo{volume}{14}, \bibinfo{number}{4} (\bibinfo{year}{1973}),
  \bibinfo{pages}{534--543}.
\newblock


\bibitem[\protect\citeauthoryear{Skov}{Skov}{2006}]%
        {skov2006role}
\bibfield{author}{\bibinfo{person}{Lise Skov}.}
  \bibinfo{year}{2006}\natexlab{}.
\newblock \showarticletitle{The role of trade fairs in the global fashion
  business}.
\newblock \bibinfo{journal}{{\em Current Sociology\/}} \bibinfo{volume}{54},
  \bibinfo{number}{5} (\bibinfo{year}{2006}), \bibinfo{pages}{764--783}.
\newblock


\bibitem[\protect\citeauthoryear{Sproles}{Sproles}{1981}]%
        {sproles1981analyzing}
\bibfield{author}{\bibinfo{person}{George~B Sproles}.}
  \bibinfo{year}{1981}\natexlab{}.
\newblock \showarticletitle{Analyzing fashion life cycles—Principles and
  perspectives}.
\newblock \bibinfo{journal}{{\em Journal of Marketing\/}} \bibinfo{volume}{45},
  \bibinfo{number}{4} (\bibinfo{year}{1981}), \bibinfo{pages}{116--124}.
\newblock


\bibitem[\protect\citeauthoryear{Summers}{Summers}{1970}]%
        {summers1970identity}
\bibfield{author}{\bibinfo{person}{John~O Summers}.}
  \bibinfo{year}{1970}\natexlab{}.
\newblock \showarticletitle{The identity of women's clothing fashion opinion
  leaders}.
\newblock \bibinfo{journal}{{\em Journal of Marketing Research\/}}
  \bibinfo{volume}{7}, \bibinfo{number}{2} (\bibinfo{year}{1970}),
  \bibinfo{pages}{178--185}.
\newblock


\bibitem[\protect\citeauthoryear{Tigert, Ring, and King}{Tigert
  et~al\mbox{.}}{1976}]%
        {tigert1976fashion}
\bibfield{author}{\bibinfo{person}{Douglas~J Tigert},
  \bibinfo{person}{Lawrence~J Ring}, {and} \bibinfo{person}{Charles~W King}.}
  \bibinfo{year}{1976}\natexlab{}.
\newblock \showarticletitle{Fashion involvement and buying behavior: A
  methodological study}.
\newblock \bibinfo{journal}{{\em ACR North American Advances\/}}
  (\bibinfo{year}{1976}).
\newblock


\bibitem[\protect\citeauthoryear{Vittayakorn, Yamaguchi, Berg, and
  Berg}{Vittayakorn et~al\mbox{.}}{2015}]%
        {vittayakorn2015runway}
\bibfield{author}{\bibinfo{person}{Sirion Vittayakorn}, \bibinfo{person}{Kota
  Yamaguchi}, \bibinfo{person}{Alexander~C Berg}, {and}
  \bibinfo{person}{Tamara~L Berg}.} \bibinfo{year}{2015}\natexlab{}.
\newblock \showarticletitle{Runway to realway: Visual analysis of fashion}. In
  \bibinfo{booktitle}{{\em 2015 IEEE Winter Conference on Applications of
  Computer Vision}}. IEEE, \bibinfo{pages}{951--958}.
\newblock


\end{thebibliography}

\end{document}